# LimeSoDa: A dataset collection for benchmarking of machine learning regressors in digital soil mapping

Jonas Schmidinger[a,b,*], Sebastian Vogel[b], Viacheslav Barkov[a,b], Anh-Duy Pham[a,b], Robin Gebbers[b], Hamed Tavakoli[b], Jose Correa[b], Tiago R. Tavares[c], Patrick Filippi[d], Edward J. Jones[d], Vojtech Lukas[e], Eric Boenecke[f], Joerg Ruehlmann[f], Ingmar Schroeter[g], Eckart Kramer[g], Stefan Paetzold[h], Masakazu Kodaira[i], Alexandre M.J.-C. Wadoux[j], Luca Bragazza[k], Konrad Metzger[k], Jingyi Huang[l], Domingos S.M. Valente[m], Jose L. Safanelli[n], Eduardo L. Bottega[o], Ricardo S.D. Dalmolin[p], Csilla Farkas[q], Alexander Steiger[r], Taciara Z. Horst[s], Leonardo Ramirez-Lopez[t,u], Thomas Scholten[v,w], Felix Stumpf[x], Pablo Rosso[y], Marcelo M. Costa[z], Rodrigo S. Zandonadi[aa], Johanna Wetterlind[ab], Martin Atzmueller[a,ac]

[a] Osnabrück University, Joint Lab Artificial Intelligence and Data Science, Osnabrück, Germany
[b] Leibniz Institute for Agricultural Engineering and Bioeconomy (ATB), Department of Agromechatronics, Potsdam, Germany
[c] University of São Paulo (USP), Center of Nuclear Energy in Agriculture (CENA), Piracicaba, Brazil
[d] The University of Sydney, Sydney Institute of Agriculture, Sydney, Australia
[e] Mendel University in Brno, Department of Agrosystems and Bioclimatology, Brno, Czech Republic
[f] Leibniz Institute of Vegetable and Ornamental Crops, Next Generation Horticultural Systems, Grossbeeren, Germany
[g] Eberswalde University for Sustainable Development, Landscape Management and Nature Conservation, Eberswalde, Germany
[h] University of Bonn, Institute of Crop Science and Resource Conservation (INRES)—Soil Science and Soil Ecology, Bonn, Germany
[i] Tokyo University of Agriculture and Technology, Institute of Agriculture, Tokyo, Japan
[j] LISAH, Univ. Montpellier, AgroParisTech, INRAE, IRD, L'Institut Agro, Montpellier, France
[k] Agroscope, Field-Crop Systems and Plant Nutrition, Nyon, Switzerland
[l] University of Wisconsin-Madison, Department of Soil Science, Madison, USA
[m] Federal University of Viçosa, Department of Agricultural Engineering, Viçosa, Brazil
[n] Woodwell Climate Research Center, Falmouth, USA
[o] Federal University of Santa Maria (UFSM), Academic Coordination, Santa Maria, Brazil
[p] Federal University of Santa Maria (UFSM), Soil Department, Santa Maria, Brazil
[q] Norwegian Institute of Bioeconomy Research (NIBIO), Division of Environment and Natural Resources, Aas, Norway
[r] University of Rostock, Chair of Geodesy and Geoinformatics, Rostock, Germany
[s] Federal Technological University of Paraná, Dois Vizinhos, Brazil
[t] BÜCHI Labortechnik AG, Data Science Department, Flawil, Switzerland
[u] Imperial College London, Imperial College Business School, London, UK
[v] University of Tübingen, Department of Geosciences, Tübingen, Germany
[w] University of Tübingen, DFG Cluster of Excellence 'Machine Learning for Science', Germany
[x] Bern University of Applied Sciences, Competence Center for Soils, Zollikofen, Switzerland
[y] Leibniz Centre for Agricultural Landscape Research (ZALF), Simulation and Data Science, Müncheberg, Germany
[z] Federal University of Jataí, Institute of Agricultural Sciences, Jatai, Brazil
[aa] Federal University of Mato Grosso, Instute of Agricultural and Environmental Sciences, Sinop, Brazil
[ab] Swedish University of Agricultural Sciences (SLU), Department of Soil and Environment, Skara, Sweden
[ac] German Research Center for Artificial Intelligence (DFKI), Research Department Plan-Based Robot Control, Osnabrück, Germany

ARTICLE INFO

ABSTRACT

Digital soil mapping (DSM) relies on a broad pool of statistical methods, yet determining the optimal method for a given context remains challenging and contentious. Benchmarking studies on multiple datasets are needed to reveal strengths and limitations of commonly used methods. Existing DSM studies usually rely on a single dataset with restricted access, leading to incomplete and potentially misleading conclusions. To address these issues, we introduce an open-access dataset collection called Precision Liming Soil Datasets (LimeSoDa). LimeSoDa consists of 31 field- and farm-scale datasets from various countries. Each dataset has three target soil properties: (1) soil organic matter or soil organic carbon, (2) clay content and (3) pH, alongside a set of features. Features are dataset-specific and were obtained by optical spectroscopy, proximal- and remote soil sensing. All datasets were aligned to a tabular format and are ready-to-use for modeling. We demonstrated the use of LimeSoDa for benchmarking by comparing the predictive performance of four learning algorithms across all datasets. This comparison included multiple linear regression (MLR), support vector regression (SVR), categorical boosting (CatBoost) and random forest (RF). The results showed that although no single algorithm was universally superior, certain algorithms performed better in specific contexts. MLR and SVR performed better on high-dimensional spectral datasets, likely due to better compatibility with principal components. In contrast, CatBoost and RF exhibited considerably better performances when applied to datasets with a moderate number (<20) of features. These benchmarking results illustrate that the performance of statistical methods can be highly context-dependent. LimeSoDa therefore provides an important resource for improving the development and evaluation of statistical methods in DSM.

* Corresponding author.
*E-mail address:* Jonas.Schmidinger@uni-osnabrueck.de (J. Schmidinger).

## 1. Introduction

In digital soil mapping (DSM), statistical modeling of the relationships between laboratory measured soil properties and secondary features is used to create soil maps for environmental and agricultural applications. A key goal of DSM research is to continuously improve the accuracy of soil maps (Minasny and McBratney, 2016). Given the statistical and computational nature of the task, most of the recent research has focused on improving accuracy by implementing more sophisticated modeling approaches—especially, since the adoption of machine learning (ML) and deep learning algorithms has increased predictive capabilities (Heuvelink and Webster, 2022). However, modeling pipelines in DSM are usually not straightforward because several statistical challenges have to be addressed such as high-dimensional, noisy and intercorrelated features (Shi et al., 2023); the incorporation of spatial and sometimes temporal components (Heuvelink and Webster, 2022); or very scarce training data (Schmidinger et al., 2024b). Various statistical methods have been suggested and developed in response. These range from different ML algorithms (Wadoux et al., 2020), pre-processing techniques (Shi et al., 2023) and data fusion approaches (Wang et al., 2022) to the development of new sampling designs (Brus, 2019). While each of the suggested methods may serve a specific purpose, it is difficult to keep track of the broad and ever-evolving pool of available tools. In addition, different studies often show contradictory conclusions about the effectiveness of commonly used methods, as discussed by Žížala et al. (2024) for sampling designs, Shi et al. (2023) for pre-processing techniques, or Heung et al. (2016) for learning algorithms. As a result, it often remains unclear how robust each of these methods is and for which context they are most suitable. Trustworthy and informative benchmarking studies are needed to fully understand their capabilities and strengths.

Benchmarking refers to the methodological comparison of competing statistical methods to assess and rank their capabilities (Nießl et al., 2022). Essentially, in a benchmarking study, the performance of two or more methods (e.g., learning algorithms) are compared based on evaluation metrics (e.g., $R^2$). Such studies are already common in DSM, as shown by our supporting literature review on benchmarking studies in 2023, provided in Appendix A. However, the informative value of them is often limited because the underlying study designs do not support comprehensive and generalizable conclusions. In the following, we will refer to the main findings from the literature review on DSM benchmarking (Appendix A) to highlight currently present shortcomings according to several guidelines on benchmarking (e.g., Boulesteix et al., 2017; Nießl et al., 2022; Weber et al., 2019).

We observed that over 95 % of DSM studies relied on a single dataset for their benchmarking **(see** Appendix A**,** Fig. A2**)**. These single datasets were either study-specific proprietary datasets or one of the few available open-access options, such as the Land Use and Coverage Area Frame Survey (LUCAS) (Orgiazzi et al., 2018). However, given that the performance of methods strongly depend on the context and inherent patterns within datasets, benchmarking should include a substantially larger number of datasets (Boulesteix et al., 2015; Strobl and Leisch, 2024). The largest number of datasets used for DSM benchmarking in 2023 was three (Appendix A**,** Fig. A2). In contrast, classical ML studies may feature tens to over a hundred datasets for their benchmarking (Shmuel et al., 2024). Including a larger number of datasets ensures a more robust analysis, as results will be less influenced by individual dataset-specific patterns. Moreover, it allows the evaluation of methods under varying conditions, which helps to uncover their context-specific strengths and limitations (Grinsztajn et al., 2022; Shmuel et al., 2024).

Another critical aspect of robust benchmarking is the open accessibility of both code and data. Fewer than 10 % of DSM benchmarking studies provided open datasets and even less than 5 % shared their code **(see** Appendix A**,** Fig. A1**)**, which is marginal compared to other computational scientific fields (Laurinavichyute et al., 2022; Pineau et al., 2021). While many studies (48 %) included a statement that indicated a willingness *'to share data upon request'*, prior studies on reproducibility have demonstrated that only a minority of researchers eventually comply with this commitment (Kratz and Strasser, 2014; Laurinavichyute et al., 2022). Therefore, most DSM benchmark studies must be deemed irreproducible. Reproducibility in the context of modeling is crucial not only for identifying potential errors through code review (Greene et al., 2017) but also because it allows researchers to build upon existing work for cumulative scientific progress (Kapoor and Narayanan, 2022). Yet, reproducibility becomes even more critical in the context of benchmarking. Benchmarking is often conducted alongside the introduction of new methods to prove their merits. It has been argued that this can make the benchmarking inherently biased towards the newly proposed method (Nießl et al., 2022; Weber et al., 2019). This is because authors can have a competitive driven interest in publishing results that show their novel method as superior to common competitors (Brown et al., 2017). Such superiority may be achieved through unfair study design choices, although often unintentionally (Nießl et al., 2022). In contrast, reproducible research with shared code and shared data encourages more rigorous and less biased study designs.

We attribute the two aforementioned shortcomings (i.e., the reliance on too few datasets in benchmarking and the lack of reproducibility), in particular to the lack of open datasets for DSM purposes. This issue has been addressed in other academic fields with the establishment of benchmark dataset collections (e.g., Morris et al., 2020; Romano et al., 2022). However, a comparable collection does not yet exist for DSM. While progress has been made to improve the data-availability as indicated by the open soil spectroscopy library (OSSL) by Safanelli et al. (2025) and other soil-databases (Gobezie et al., 2024), the few existing open soil datasets are primarily focused on large-scale laboratory-based soil spectroscopy or include only the target soil properties without accompanying predictor features. Furthermore, some open datasets (e.g., LUCAS) were published under restrictive licenses, which prohibits sharing of the dataset within a code repository. On the other hand, open datasets from a field- and farm-scale for high-resolution precision agriculture are almost completely missing, despite the fact that precision agriculture with proximal soil sensors is an important application of DSM (Gebbers and Adamchuk, 2010). Field-scale DSM with proximal soil sensors has its own distinctive challenges such as training data scarcity (Schmidinger et al., 2024b) next to high-dimensional data fusion (Schmidinger et al., 2024a; Wang et al., 2022). To address these issues, we introduce *Precision Liming Soil Datasets* (LimeSoDa). LimeSoDa is an open dataset collection with small-scale datasets, that are ready-to-use for modeling. This expands the possibility of DSM practitioners to benchmark statistical methods across a variety of soil mapping contexts. In **Section 2**, we provide an overview of the datasets included in LimeSoDa. Additionally, we present an example benchmark study in which we compared four learning algorithms (**Section 3**), to demonstrate how LimeSoDa can lead to more comprehensive and nuanced conclusions. Lastly, **Section 4** offers an outlook on the potential applications of LimeSoDa beyond the present study.

## 2. LimeSoDa

### 2.1. Overview

LimeSoDa contains 31 datasets at field- to farm-scale. Datasets are ready-to-use for modeling, which means that they contain continuous target soil properties and features in a tabular format, usable for regression tasks. Three target soil properties are included in all datasets: soil organic matter (SOM) or soil organic carbon (SOC), pH and clay. Since SOC and SOM are directly related, we treated them interchangeably as a single soil property. These three soil properties are among the most commonly evaluated soil properties in DSM (Chen et al., 2022) and are crucial parameters for assessing soil quality. We refer to this benchmarking collection as “LimeSoDa” because these three target soil properties are not exclusively but especially relevant for lime requirement calculations, according to best management practices in the UK and Germany (Agriculture Horticulture Development Board, 2023; Bönecke et al., 2021). Features for modeling are dataset-specific and originate from laboratory-based spectroscopy, in-situ proximal soil sensing, and remote sensing. Datasets are released under a permissive open-access license, allowing implementation in a code-repository to increase the possibility of code sharing. Additionally, pre-determined folds for cross validation (CV) are provided for each dataset, to enable comparability with our benchmarking and future benchmarking results. In total, the combined datasets include 3,174 soil samples but sample sizes of individual datasets range from 30 to 460 samples.

Pre-processing is not mandatory when working with LimeSoDa datasets. However, for datasets with high-dimensional spectral data, feature dimensionality reduction is strongly encouraged. Nevertheless, further pre-processing may improve modeling performances, an aspect that should itself be benchmarked using this dataset collection.

**Table 1**
List of datasets included in LimeSoDa.

| Dataset ID | Location | Study Area (ha) | Number of Samples | Number of Features | Sensor Data* | Previous Usage** |
|---|---|---|---|---|---|---|
| SSP.460 | State of Sao Paulo, Brazil | 473 | 460 | 830 | vis-NIR | Ramirez-Lopez et al. (2019) |
| BB.250 | Brandenburg, Germany | 52 | 250 | 17 | DEM, ERa, Gamma, pH-ISE, RSS, VI | Schmidinger et al. (2024b) |
| SP.231 | Saitama Prefecture, Japan | 3.1 | 231 | 272 | vis-NIR | Kodaira and Shibusawa (2020) |
| B.204 | Bahia, Brazil | 204 | 204 | 16 | DEM, RSS, VI | Pereira et al. (2022) |
| G.150 | Goias, Brazil | 79 | 150 | 17 | DEM, ERa, RSS, VI | Valente et al. (2024) |
| H.138 | Hubei, China | 420 | 138 | 2,489 | MIR | Wadoux et al. (2025) |
| SL.125 | Skåne Län, Sweden | 78 | 125 | 2,082 | ERa, vis-NIR | Wetterlind et al. (2010) |
| UL.120 | Uppsala Län, Sweden | 97 | 120 | 2,082 | ERa, vis-NIR | Wetterlind et al. (2010) |
| NRW.115 | North Rhine-Westphalia, Germany | 17 | 115 | 1,686 | MIR | Leenen et al. (2022) |
| MG.112 | Mato Grosso, Brazil | 111 | 112 | 17 | DEM, ERa, RSS, VI | Valente et al. (2024) |
| SA.112 | Saxony-Anhalt, Germany | 27 | 112 | 1,412 | DEM, ERa, Gamma, NIR, pH-ISE, VI | – |
| G.104 | Goias, Brazil | 95 | 104 | 16 | DEM, RSS, VI | – |
| MGS.101 | Mato Grosso do Sul, Brazil | 95 | 101 | 16 | DEM, RSS, VI | – |
| CV.98 | Canton of Vaud, Switzerland | 28 | 98 | 2,151 | vis-NIR | Metzger et al. (2024) |
| SC.93 | Santa CatariMG-na, Brazil | 108 | 93 | 2,146 | vis-NIR | Horst et al. (2018) |
| BB.72 | Brandenburg, Germany | 3.4 | 72 | 17 | DEM, ERa, Gamma, pH-ISE, RSS, VI | – |
| NRW.62 | North Rhine-Westphalia, Germany | 0.6 | 62 | 1,686 | MIR | Leenen et al. (2019) |
| RP.62 | Rhineland-Palatinate, Germany | 3.3 | 62 | 1,410 | ERa, Gamma, NIR, pH-ISE, VI | Tavakoli et al. (2022) |
| SSP.58 | State of Sao Paulo, Brazil | 0.7 | 58 | 351 | vis-NIR | Tavares et al. (2020) |
| NSW.52 | New South Wales, Australia | 1,158 | 52 | 5 | DEM, RSS | Filippi et al. (2019)Jones et al. (2021) |
| BB.51 | Brandenburg, Germany | 40 | 51 | 4 | DEM, ERa, pH-ISE | – |
| SC.50 | Santa Catarina, Brazil | 13 | 50 | 3 | DEM, ERa | Bottega et al. (2022) |
| W.50 | Wisconsin, USA | 80 | 50 | 15 | DEM, ERa, VI, XRF | Chatterjee et al. (2021) |
| PC.45 | Pest County, Hungary | 4.5 | 45 | 4 | CSMoisture, ERa | Ristolainen et al. (2006) |
| MG.44 | Mato Grosso, Brazil | 13 | 44 | 351 | vis-NIR | Tavares et al. (2020) |
| NRW.42 | North Rhine-Westphalia, Germany | 1.5 | 42 | 1,686 | MIR | Leenen et al. (2019) |
| SM.40 | South Moravia, Czechia | 53 | 40 | 3 | DEM, ERa | Lukas et al. (2009) |
| MWP.36 | Mecklenburg-Western Pomerania, Germany | 18 | 36 | 5 | DEM, RSS | Steiger et al. (2025) |
| O.32 | Occitania, France | 1.5 | 32 | 1,637 | MIR | Wehrle et al. (2022) |
| BB.30_1 | Brandenburg, Germany | 19 | 30 | 8 | DEM, ERa, pH-ISE, VI | – |
| BB.30_2 | Brandenburg, Germany | 1.4 | 30 | 13 | DEM, ERa, Gamma, RSS, VI | – |

*Abbreviations: Capacitive soil moisture (CSMoisture), Digital elevation model and terrain parameters (DEM); Apparent electrical resistivity (ERa); Gamma-ray activity (Gamma); Mid infrared spectroscopy (MIR); Near infrared spectroscopy (NIR); Ion selective electrodes for pH determination (pH-ISE); Remote sensing derived spectral data (RSS), X-ray fluorescence derived elemental concentrations (XRF), Vegetation indices (VI), Visible- and near infrared spectroscopy (vis-NIR).
** The datasets used in the referenced studies does not always completely align with the datasets of LimeSoDa, because only a subset or additional data may have been used.

All datasets were collected in the context of previous studies or research projects. The vast majority of these datasets had not been publicly available prior to this effort. Hence, LimeSoDa is built upon voluntarily submissions from researchers and research institutes. An overview of each dataset is provided in Table 1. Datasets were included based on the following four criteria:

**Precision liming context:** A dataset had to contain SOM/SOC, pH, and clay as topsoil (<30 cm) target properties at field- and farm-scale (<2,000 ha).

**Wet chemistry:** Target soil properties had to be determined through wet chemistry techniques instead of being inferred from spectral models.

**Predictive performance:** Target soil properties had to be predictable with the given dataset-specific set of features. We excluded datasets where no learning algorithm outperformed the null model (i.e., with an $R^2 < 0$) for any of the three target soil properties given the modeling pipeline in **Section 3.1**.

**Sample size:** A dataset had to contain a minimum size of at least 30 soil samples for all three soil properties at the same distinct sampling locations.

Datasets were not excluded because of their sampling design. For example, some datasets were based on a targeted sampling design (see Appendix B, Table B1), which refers to sampling focused on specific areas or features of interest rather than random or systematic sampling. The validation of soil maps generated from targeted- or spatially clustered sampling without an additional independent testing sample set is controversial because the absolute performance may not be reliably estimated through CV or data splitting (Piikki et al., 2021). However, we argue that for the purpose of benchmarking, the relative performance differences remain informative. Nonetheless, users of LimeSoDa can decide if they prefer other validation strategies that take into account the spatial dependency or pattern of a dataset.

Spatial coordinates are available for more than half of the datasets (see Appendix B, Fig. B1). In other cases, coordinates had to be excluded or anonymized mostly because of privacy concerns.

### 2.2. Processing of datasets

Only a minimal amount of processing was conducted for datasets of LimeSoDa to ensure that the data was as close to the original form as possible while still enabling effective benchmarking. This gives users of LimeSoDa the flexibility to apply their own pre-processing on the datasets. Ordinary kriging was employed for datasets collected with on-the-go proximal soil sensors to match sensing locations with soil sampling locations. For features available in raster format (DEM, RSS and VI), feature values were extracted at the soil sampling locations. Some sensors had an underlying data-processing step within their internal software. This led to some optical spectroscopy data being resampled to different wavebands and spectral resolutions compared to the measured raw data. For datasets with limited training samples (< 60 samples), we included fewer remote sensing-based features to maintain a favorable sample-to-feature ratio. Processing steps are documented in more detail in the metadata of the datasets. Samples with missing values in either the feature or target soil property matrix were always discarded. Two spectroscopy datasets (H.138 and NRW.115) had a few samples with reflectance values above 100 % for certain bands (see Appendix B, Fig. B2). Whether these anomalies are due to noise or instrumental calibration is unclear. Nevertheless, we did not discard these samples or bands from the dataset.

It is important to note that datasets across LimeSoDa were deliberately not harmonized. While this reduces the possibility of cross-dataset learning, it increased the flexibility of including datasets from various contexts and domains without relying on many assumptions necessary for data harmonization. The aim of LimeSoDa is not to build a unified database but to provide multiple smaller datasets that are independent entries for a benchmarking study. As a consequence, target soil properties are sometimes expressed in different units depending on the measurement method (see Appendix B, Fig. B2). This distinguishes LimeSoDa from spectral libraries, that rely on harmonization of spectral features and soil properties (Safanelli et al., 2025). Nonetheless, for coherence within LimeSoDa, we transformed the apparent electrical conductivity to ERa and spectral measurements are expressed as reflectance.

### 2.3. Accessibility

LimeSoDa is closely aligned to the FAIR-principles (Wilkinson et al., 2016). It is freely accessible through Zenodo (https://doi.org/10.5281/zenodo.14932573), is licensed under CC BY-SA 4.0 and contains extensive documentation in the form of dataset-specific metadata. Additionally, an R- and Python dataset package, called likewise *LimeSoDa* downloadable from GitHub, was created. It can be accessed through github.com/JonasSchmidinger/LimeSoDa for R and https://github.com/a11to1n3/LimeSoDa for Python.

## 3. Demonstrative benchmark study

### 3.1. Methodology

A benchmarking study was conducted to demonstrate the use of LimeSoDa. Four learning algorithms were compared based on their predictive performance. These included random forest (RF) (Breiman, 2001) and support vector regression (SVR) due to their widespread use in DSM (Khaledian and Miller, 2020), categorical boosting (CatBoost) (Prokhorenkova et al., 2018) because of its strong performances in recent ML benchmarking studies on tabular datasets (McElfresh et al., 2023; Shmuel et al., 2024) and multiple linear regression (MLR) to include a simple baseline model. We trained and evaluated each algorithm independently on SOC/SOM, pH and clay for each of the 31 datasets of LimeSoDa. Consequently, a total of 93 prediction tasks (31 datasets × 3 target soil properties) were utilized in the benchmarking. Across these 93 prediction tasks, algorithms were compared on the $R^2$ and the ordinally ranked root mean square error (RMSE) **(see** Appendix C.1). Additionally, the Wilcoxon Signed-Rank Test was used to determine whether the mean ordinal ranks based on the RMSE significantly differed ($\alpha = 0.1$) among the learning algorithms.

The learning algorithms were evaluated and hyperparameters tuned, using a nested $K$-fold cross validation (CV), with the outer loop ($K = 10$) used for model testing and the inner loop ($K' = 5$) for hyperparameter selection. The optimal hyperparameter combinations for SVR, CatBoost and RF were selected through a random search with 400 iterations based on the lowest aggregated RMSE of the inner loop. The hyperparameter search-space is given in Table 2**.**

For datasets with high-dimensional optical spectroscopy data (vis-NIR, NIR or MIR), modeling with the raw data is strongly limited due to the unfavorable sample to feature ratio. Hence, we duplicated the hyperparameter search space to 800 iterations and added two unsupervised but computationally efficient methods for dimensionality reduction as additional hyperparameter branch: principal component analysis (PCA) and a correlation matrix filter (CMF) (Perez-Riverol et al., 2017). CMF and PCA were applied only to the vis-NIR, NIR, or MIR features after the separation of training and testing or validation folds. With CMF, spectral features were discarded when the absolute pairwise Pearson correlation coefficient exceeded a defined cutoff, following the *findCorrelation* algorithm of the R-package *caret* (Kuhn, 2008). The cutoff for CMF ranged from 0.7 to 1, where CMF = 1 refers to no dimensionality reduction, and the number of searched principal components for PCA from 5 to 20. Algorithm 1 presents the modeling pipeline as pseudocode. The actual R-code is available on github.com/JonasSchmidinger/LimeSoDa_benchmarking.

**Table 2**
Hyperparameter search-space for SVR, CatBoost and RF, where a random instance is drawn from the given distribution. Names of the hyperparameters refer to the ranger R-package (Wright and Ziegler, 2017) for RF, catboost R-package for CatBoost (Dmitriev et al., 2024) and e1071 R-package (Meyer et al., 2024) for SVR.

| Learning Algorithm | Hyperparameter | Search Space | Distribution |
|---|---|---|---|
| CatBoost | depth | [1, 10] | Discrete Uniform |
| | learning_rate | [0.005, 0.5] | Log-Uniform |
| | iterations | [50, 2000] | Discrete Uniform |
| | l2_leaf_reg | [0, 10] | Uniform |
| | rsm | [0.6, 1] | Uniform |
| | subsample | [0.6, 1] | Uniform |
| | random_strength | [0.001, 10] | Log-Uniform |
| RF | num.trees | = 2 000 | – |
| | mtry | [0.1, 1]* | Uniform |
| | min.node.size | [1, 12] | Discrete Uniform |
| | max.depth | [1, 10] | Discrete Uniform |
| | sample.fraction | [0.6, 1] | Uniform |
| SVR | cost | [0.01, 1000] | Log-Uniform |
| | gamma | [0.001, 10] | Log-Uniform |
| | kernel | [linear, radial] | Discrete Uniform |

* Given as fraction instead of absolute value input for R-function.

### 3.2. Results & Discussion

The $R^2$ distribution obtained from all prediction tasks with the four different learning algorithms is shown in Fig. 1**a**. Overall, there appear to be small differences in the performance distribution. Although SVR and CatBoost achieved the best average $R^2$ of 0.44 across all prediction tasks, they were only moderately better than MLR, which had the lowest average $R^2$ of 0.40. Similarly, this is also reflected by the ordinally ranked RMSE in Fig. 2**a**. It illustrates how frequently a prediction algorithm achieved each rank compared to others, based on the ascending RMSE. For example, MLR had the lowest (i.e., the best) RMSE for 27 % of the prediction tasks, receiving the first rank. Simultaneously, it had the highest (i.e., the worst) RMSE for 30 % of prediction tasks, placing it in the fourth rank. Overall, Fig. 2**a** resembles a uniform-like distribution. This means that each learning algorithm had roughly the same number of prediction tasks in which they ranked best to worst. The Wilcoxon Signed-Rank Test further confirmed that differences among the learning algorithms were not statistically significant **(**Fig. 2**a)**. This might be surprising to some DSM practitioners, because more sophisticated methods are often expected to significantly outperform a simple model like MLR (Padarian et al., 2020). However, different learning algorithms, including simplistic models like MLR, have advantages under certain conditions, as illustrated in the next paragraphs.

Grouping the prediction tasks based on the presence or absence of vis-NIR, NIR or MIR features reveals a significant shift in the performance distribution among the learning algorithms **(**Fig. 1**b-c &** Fig. 2**b-c)**. The grouping is made because datasets with vis-NIR, NIR and MIR are characterized by a highly inflated feature to training sample ratio, so that dimensionality reduction is usually necessary prior to the modeling. On datasets without vis-NIR, NIR and MIR, tree-based learning

**Algorithm 1.** Pseudocode of the modeling pipeline used in the benchmarking

```
Input
D ← list of 31 datasets
S ← list of 3 target soil properties
H ← list of 400 hyperparameter combinations
for d in D do:
  if vis-NIR, NIR or MIR features in d do:
    replicate H with PCA and CMF as additional hyperparameter
  split d into K = 10 outer folds
  for s in S do:
    for k = 1 to K do:
      k ← testing fold
      K – 1 ←outer training dataset
      split K – 1 into K' = 5 inner folds
      for h in H do:
        for k' = 1 to K' do:
          k' ← validation fold
          K' – 1 ←inner training dataset
          train CatBoost, SVR, MLR* and RF on s with K'– 1 given h
          predict s in k' with trained CatBoost, SVR, MLR* and RF
        compile s in k' predictions and observations
        compute RMSE from compiled s in k' predictions and observations
      select h with lowest RMSE (h^)
      train CatBoost, SVR, MLR and RF on s with K– 1 given h^
      predict s in k with trained CatBoost, SVR, MLR and RF
    compile s in k predictions and observations
    compute RMSE and R² from compiled s in k predictions and observations
store RMSE and R² of each s and d
```

* Conditional hyperparameters for dimensionality reduction searched for MLR in inner-loop only when vis-NIR, NIR or MIR features in *d*.

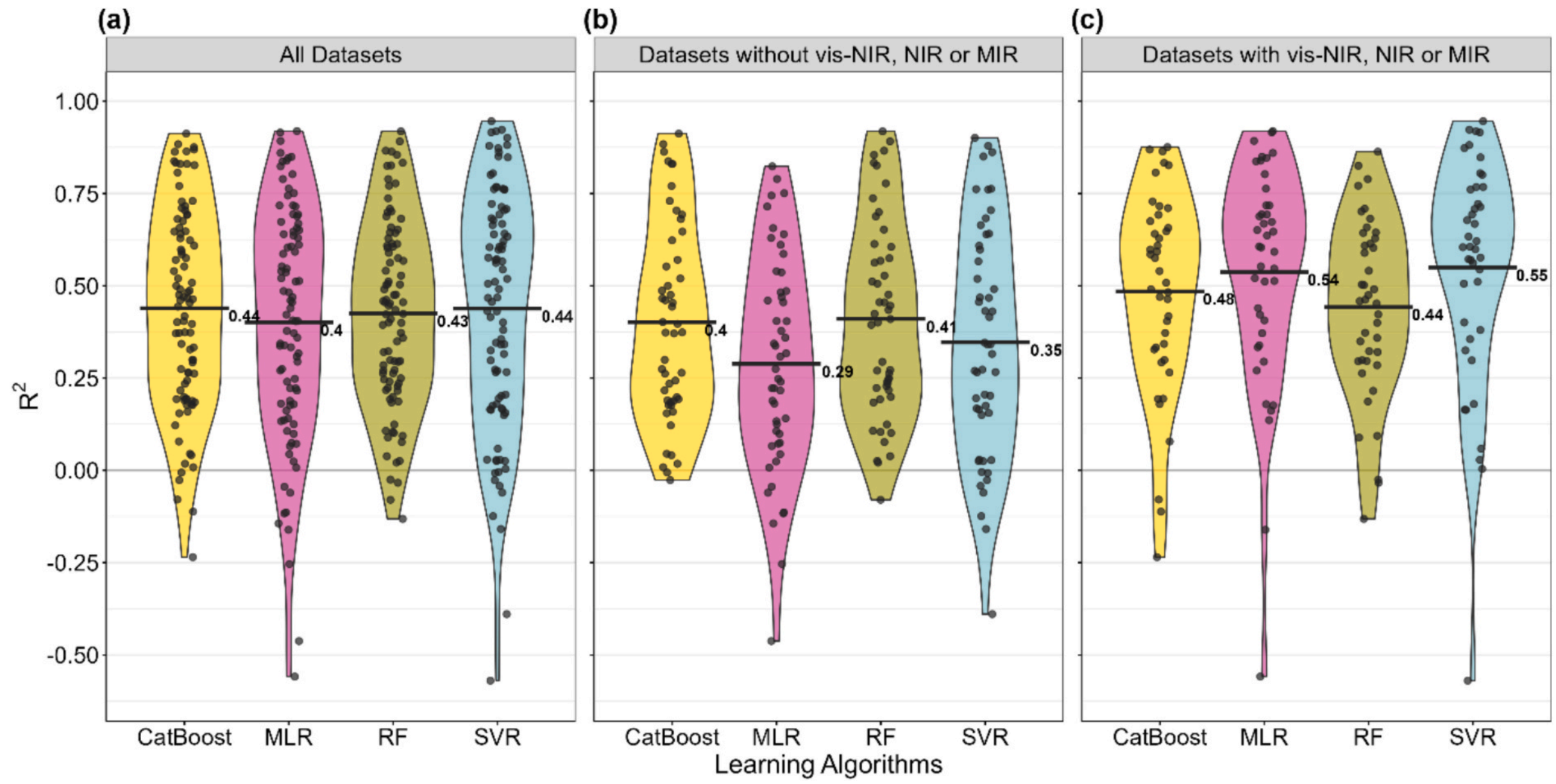

**Fig. 1.** Violin plot showing the distribution of $R^2$ values grouped by learning algorithms for prediction tasks from (a) all datasets, (b) datasets without vis-NIR, NIR, or MIR, and (c) datasets with vis-NIR, NIR, or MIR. Horizontal lines with labels represent the mean $R^2$ value.

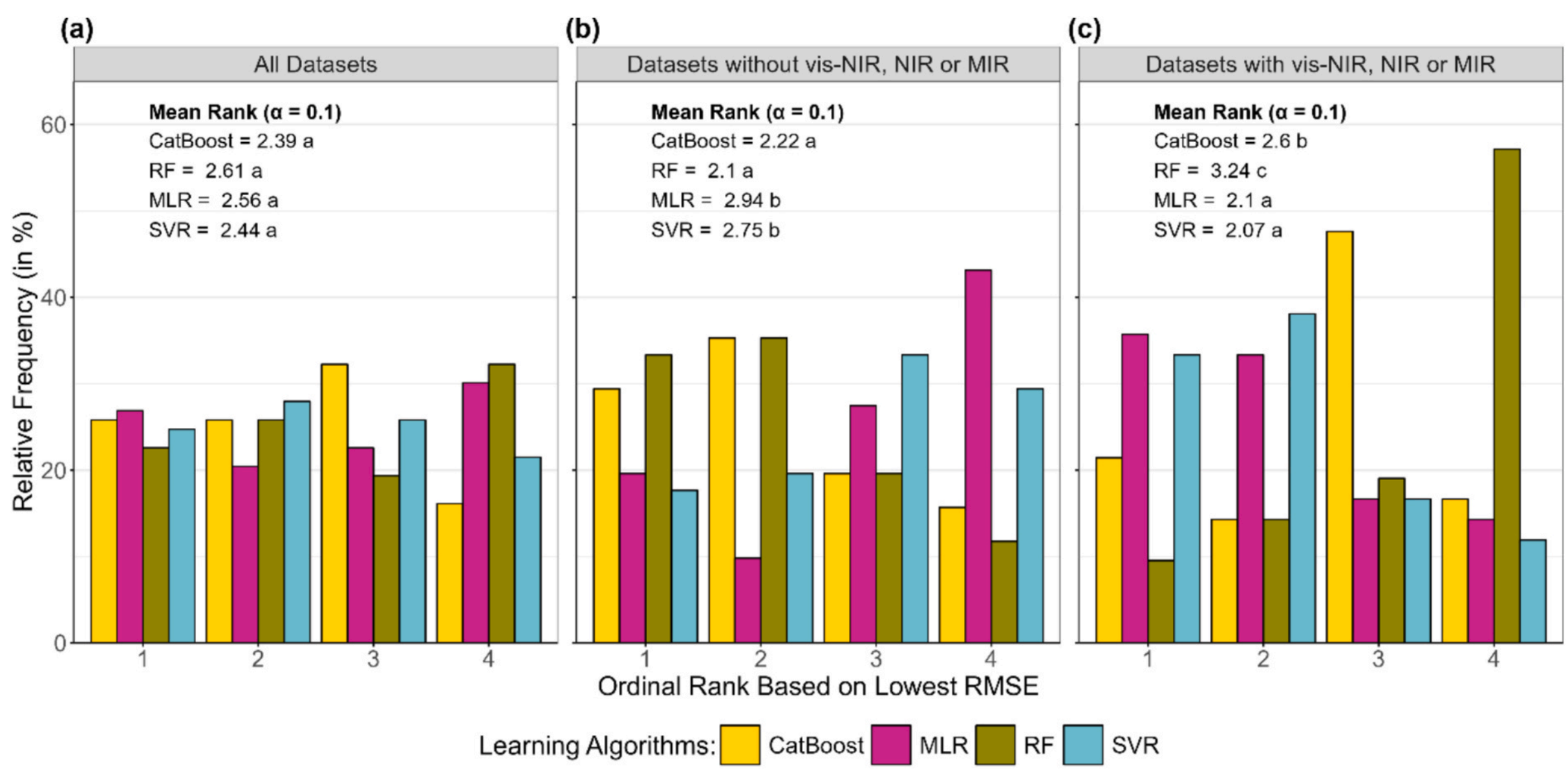

**Fig. 2.** Bar plot showing the relative frequency distribution of ordinal ranks based on the lowest RMSE (i.e., lower rank number indicates better performance) for the four learning algorithms for predictions tasks from (a) all datasets, (b) datasets without vis-NIR, NIR, or MIR, and (c) datasets with vis-NIR, NIR, or MIR. Mean ranks of the ordinally ranked RMSE for the learning algorithms are shown, including significance letters from the Wilcoxon Signed-Rank Test (α = 0.1) to indicate whether the performance differences between the learning algorithms were statistically significant.

algorithms (i.e., RF and CatBoost) considerably outperformed SVR and MLR (Fig. 1**b**). This is further highlighted by the fact that in 63 % of the prediction tasks in these datasets, RF or CatBoost had the best performance (Fig. 2**b**). Meanwhile, MLR or SVR had the worst performance in 72.5 % of the cases. The advantages of tree-based learning algorithms for regular tabular datasets have been highlighted in numerous benchmarking studies (Grinsztajn et al., 2022; Shmuel et al., 2024; Shwartz-Ziv and Armon, 2022). Their particular strengths include effective regularization when dealing with multicollinearity and irrelevant features, along with the ability to learn irregular functions or interactions. For this reason, tree-based algorithms, most notably RF, have become the most widely used learning algorithms in DSM (Khaledian and Miller, 2020).

Opposite results are evident for datasets with vis-NIR, NIR and MIR. Here, SVR and MLR had on average an $R^2$ considerably larger than that of CatBoost and RF (Fig. 1**c**). Additionally, they ranked first in 69 % of the prediction tasks (Fig. 2**c**), whereas RF or CatBoost had the worst performances in 73.8 % of the cases. The very high dimensionality and inherent multicollinearity in spectral data, where adjacent bands are often highly correlated, made feature dimensionality reduction techniques, such as PCA and CMF, essential. For example, the dataset O.32 had up to 1,637 features but only 32 soil samples available for training (Table 1). Learning algorithms would severely overfit on this inflated feature to training sample ratio. However, CMF and especially PCA facilitated proper modeling, as shown by the best selected hyperparameters (see Appendix C.2, Fig. 1C). This posed a disadvantage to the tree-based algorithms because PCA has been reported to perform weaker when combined with tree-based algorithms (Howley et al., 2005). PCA creates new uncorrelated features that are linear combinations of the original features but tree-based models are inflexible to adapt to rotated or transformed data that do not represent the original feature composition (Grinsztajn et al., 2022). Consequently, MLR and SVR outperformed CatBoost and RF due to their better compatibility with PCA. Previous benchmarking studies in soil spectroscopy, as reviewed by Padarian et al. (2020), have reported different outcomes favoring more sophisticated methods. One possible explanation for this discrepancy is that these earlier studies relied on a single dataset, which may inadvertently lead to overinterpretation of incidental results. Alternatively, an unintended publication bias favoring novel methods over simpler ones, as observed in other computational fields (Buchka et al., 2021), could have contributed to the differing outcomes. Nonetheless, we acknowledge that tree-based models possibly perform better on full spectral data when very large training sets are available (Clingensmith and Grunwald, 2022) or with effective feature selection (Canero et al., 2024). Such an extended analysis was not within the scope of this study.

Lastly, we found a relationship between the mean ordinal rank of a learning algorithm and the sample size available for a prediction task (Fig. 3). Most notably, MLR showed sensitivity to the number of available samples, as its mean rank worsened for datasets with more than 100 samples. This aligns with previous results, which similarly indicated that MLR can have advantages for small sized datasets but may not be ideal if enough training data is available (Schmidinger et al., 2024b). The opposite behavior was observed for SVR, which considerably improved its rank for datasets with more than 100 samples. For RF and CatBoost, the effect of the sample size was not as pronounced. CatBoost was slightly better with more training data, whereas RF was more effective for smaller datasets.

Despite the advantages that certain learning algorithms have, their superiority is not deterministic. Their performances can vary based on numerous factors and may not be optimal in every case. For example, RF was generally the worst performing learning algorithm for datasets with vis-NIR, NIR and MIR (Fig. 1**b**), yet it still turned out to be best in 10 % of the prediction tasks (Fig. 2**b**). Therefore, relying on a single dataset is insufficient to establish the superiority of one method and may foster misleading conclusions based on non-generalizable results.

## 4. Further applications

For demonstrative purposes, we restricted the benchmarking to four learning algorithms, although many other relevant methods remain to be explored. The benchmarking study of **Section 3.2** is fully reproducible, as the code and datasets have been published. Therefore, additional learning algorithms or different feature selection strategies can readily be integrated into the pipeline of **Section 3.1** to further extend the analysis. While neural networks were not included in this initial benchmarking due to their relatively poor performances in larger tabular benchmarking studies (Shmuel et al., 2024), recent advancements in neural networks designed for tabular data such as TabPFN (Hollmann et al., 2025) have demonstrated promising results. First successful applications of TabPFN in DSM can be found by Barkov et al. (2024) but its application has to be further evaluated. Lastly, we encourage extending the benchmarking with more open datasets, such as those available from OSSL, in addition to LimeSoDa.

Within the scope of this study, we focused entirely on the need of open datasets for benchmarking purposes to address currently present shortcomings. Nonetheless, the usage of LimeSoDa should not be restricted to statistical benchmarking or method development. Open datasets from various geographical context with different sensing techniques may enhance our general understanding of pedological processes. It allows for the critical evaluation of soil mapping in agronomical decision making and asses the robustness of various soil sensing techniques. Especially the liming context of the dataset can be useful for studies in the field of precision agriculture. Consequently, LimeSoDa is a valuable tool for addressing key challenges of pedometrics outside of DSM benchmarking (Wadoux et al., 2021). In other academic fields, secondary data analysis has already answered various research questions unrelated to the original research purpose of the dataset (Greene et al., 2017). Lastly, spectroscopy data from LimeSoDa can be harmonized and added to other spectral libraries as additional training data to improve global modeling.

## 5. Conclusion

Current benchmarking practices in DSM suffer from data limitations that lead to incomplete or potentially biased conclusions. There is a lack of open datasets from various domains, spatial dimensions and types of sensors. To address this problem, we introduced an open-access data collection called LimeSoDa, which currently consists of 31 field- and farm-scale datasets. LimeSoDa offers datasets that are ready-to-use for modeling and covers various types of features from different sensing techniques. This enables those who are working in the field of DSM to benchmark statistical methods on a diverse range of soil datasets. Further, the open license is intended to improve the reproducibility.

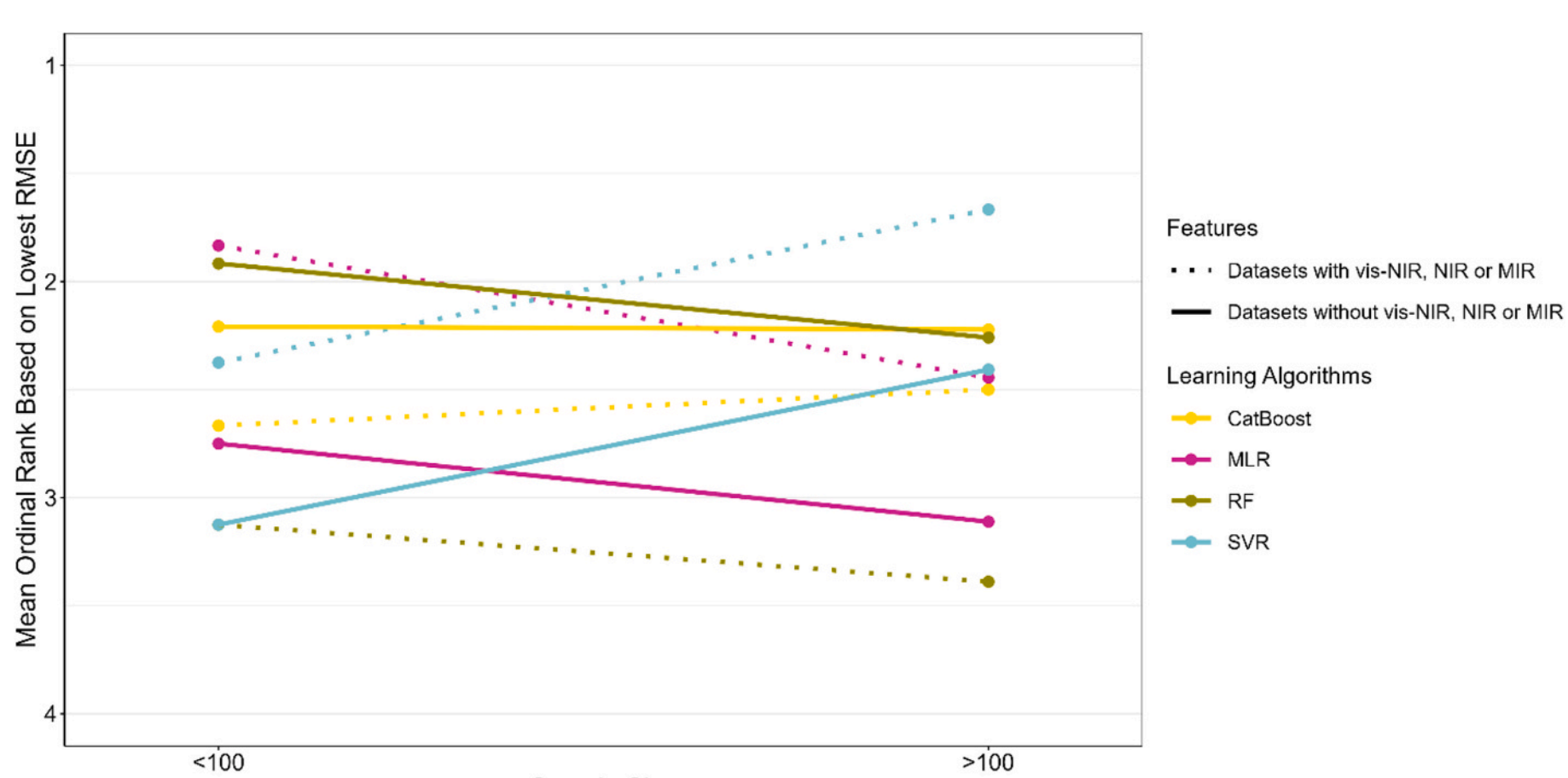

**Fig. 3.** Line plot showing the mean ordinal rank based on the lowest RMSE (i.e., lower rank number indicates better performance) in dependence to the sample size. Lines are differentiated by the four learning algorithms and the type of features present in the dataset of the prediction task.

The utility of LimeSoDa was demonstrated through a benchmarking study with four learning algorithms. The results showed that no algorithm significantly exceeded the others across all datasets. Instead, different learning algorithms had advantages depending on the type of features and sample size of a dataset. On average, tree-based algorithms, i.e., CatBoost and RF, performed better on datasets without vis-NIR, NIR and MIR features, proving their suitability for conventional tabular datasets. In contrast, SVR and MLR outperformed CatBoost and RF for datasets with vis-NIR, NIR and MIR due to their better compatibility with PCA-transformed data. Additionally, the training sample size influenced the ranking of a learning algorithm. The relative performance most notably decreased for MLR and increased for SVR with more training samples.

A benchmarking study based on a single dataset could not have revealed such context-dependent performance of learning algorithms. More so, relying on singular datasets risks overinterpreting incidental outcomes and a potential publication bias favoring newer methods cannot be ruled out. In contrast, LimeSoDa facilitates more in-depth analyses and enables comprehensive conclusions. Additionally, the benchmarking can readily be extended with further learning algorithms or pre-processing techniques because datasets and code are openly available.

Beyond benchmarking, there are further applications and challenges in pedometrics that benefit from open datasets. LimeSoDa can be used to investigate pedological processes across diverse geographical contexts or the spectral data can be integrated into other spectroscopy libraries to improve global modeling. In summary, by providing a rich and diverse collection of open datasets, LimeSoDa has the potential to significantly advance ML applications in pedometrics.

## Funding

This research was supported by the Lower Saxony Ministry of Science and Culture (MWK), funded through the zukunft.niedersachsen program of the Volkswagen Foundation (ZN4072) as well as the Federal Ministry of Education and Research of Germany (BMBF) through the BonaRes project I4S: Intelligence for Soil (031B1069A). Compute resources were funded by the Deutsche Forschungsgemeinschaft (DFG, German Research Foundation) project number 456666331.

## Declaration of competing interest

The authors declare that they have no known competing financial interests or personal relationships that could have appeared to influence the work reported in this paper.

## Acknowledgements

We are thankful to a lot of scientists, field- and technical assistants who helped us to gather the data and metadata of the datasets.

## Appendix A:. Literature review on DSM benchmarking studies

### *Appendix A.1: Literature review methodology*

The “Web of Science” database was used for the literature review. We searched for “digital soil mapping” and “predictive soil mapping” as keywords in the abstract for documents defined as “data paper”, “early access” or “article” and restricted the search to publications from the year 2023. This resulted in an initial pool of 192 papers. From this pool, eight papers were excluded because they were not written in English, missed DSM context or were inaccessible. Another 73 were removed because the abstract did not indicate that any statistical benchmarking was conducted, leading to a total of 111 papers for the review on DSM benchmarking practices. As benchmarking we considered a broad definition, encompassing any comparison of competing statistical methods based on quantified validation metrics. This included comparisons of: prediction algorithms, feature selection methods, sampling designs, engineered features etc. We did not consider the comparison of sensors or features as benchmarking, if the focus did not lie on the statistical engineering of the features (e.g., temporal stacks). In some studies, the benchmarking was not the main objective of the study but rather supplementary information. We also considered these as “benchmarking study” when the comparison was mentioned in the abstract.

The code and data availability in each of the 111 papers was evaluated. We differentiated between “not shared”, “shared” and “partially shared”. The statement on material availability and any supplementary information provided were examined alongside keywords such as “repository”, “dataset”, “GitHub”, “Zenodo”, “code” etc. to determine the availability of code or data. We classified any data or code to be “shared” when they were directly accessible through the supplementary information or a repository (e.g., Zenodo or GitHub). In some code-repositories, datasets were not directly provided due to restrictive dataset licenses (e.g., as in LUCAS) but links to the websites that hosted these datasets were added. This was also considered as “shared” when it was provided in combination with the dataset pre-processing code. In other cases, links were given but they expired since the publication or the dataset hosting webpages were not in English, which made navigation impossible. These cases were considered “not shared”. Lastly, it was evaluated how many datasets were used for the benchmarking. We classified any collection of data based on a coherent sampling in a distinct study area as dataset.

### *Appendix A.2: Literature review results*

A majority of over 90 % neither shared the data nor the code (Fig. A1). However, many studies (48 %) included a statement on the data availability with willingness to provide data on request. In contrast, such a statement on the availability of code was almost never included apart from a single exception (1 %). This can be explained by the fact that most journals ask for a statement on the availability of data but do not expect this for the code availability.

The majority of benchmarking was conducted based on a single dataset (95.5 %) and the maximum number of datasets did not exceed three (Fig. A2). All materials used in the literature review are provided at github.com/JonasSchmidinger/LimeSoDa_literature.review, with additional comments for boundary cases.

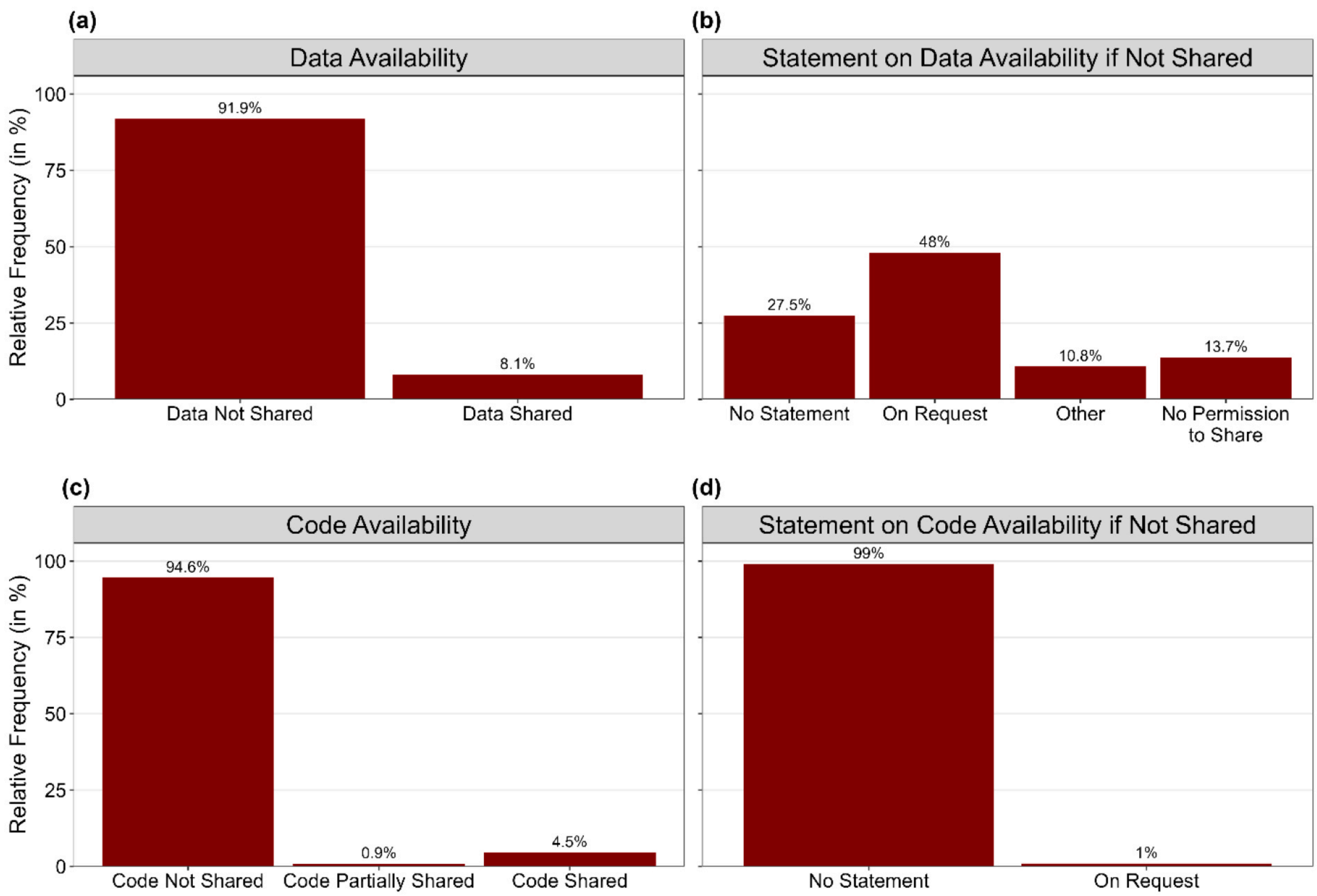

**Fig. A1.** Bar plot showing the relative frequency distribution on the availability of data (a) or code (c) in other DSM benchmarking publications from 2023, as well as the statement on data (b) or code (d) availability if datasets were not shared.

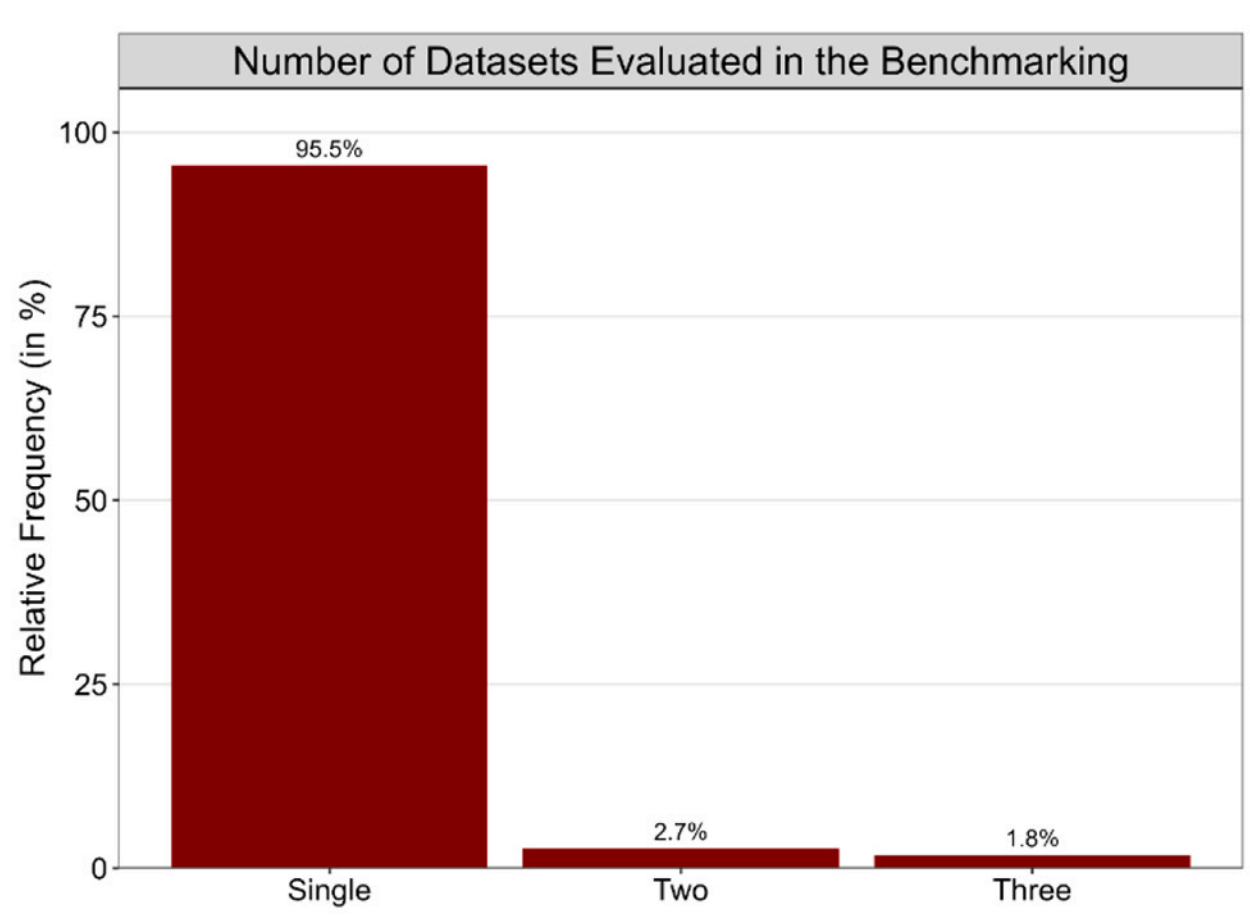

**Fig. A2.** Bar plot showing the relative frequency distribution on how many datasets were evaluated in other DSM benchmarking publications from 2023.

## Appendix B:. Further information on LimeSoDa

In the following section, we added further information and exploratory figures describing datasets of LimeSoDa. Table B1 provides information on the sampling design that was used to create the dataset and the availability of spatial coordinates. Nonetheless, coordinates may be provided on request by the authors if they are not included. The locations of the datasets are further depicted on a global map in Fig. B1. Fig. B2 illustrates the spectral reflectance curves of vis-NIR, NIR and MIR from each sample. Spectral data were provided as raw as possible in order to allow users of LimeSoDa to use their own processing methods. Fig. B3 shows the distribution of target soil property values per dataset. Since SOC and SOM were not harmonized, they are presented separately. Additionally, MG.44 and SSP.58 did not measure SOC and clay as gravimetric percentage but as concentration. Therefore, they needed an independent scale. Fig. B4 includes the correlation plots of the datasets. For concise visualization, we only selected a single feature with the highest absolute correlation to the target soil properties.

**Table B1**
Extended overview of datasets included in LimeSoDa.

| Dataset ID | Sampling Design | Availability of Spatial Coordinates |
|---|---|---|
| SSP.460 | Regular Grid Sampling | Without Coordinates |
| BB.250 | Triangular Grid Sampling | With Coordinates |
| SP.231 | Random Sampling & Systematic Sampling | With Coordinates |
| B.204 | Regular Grid Sampling | With Coordinates |
| G.150 | Regular Grid Sampling | With Coordinates |
| H.138 | Adapted Latin Hypercube Sampling & Uncertainty Guided Sampling | With Coordinates |
| SL.125 | Regular Grid Sampling & Surface Tortoise Sampling | With Dummy Coordinates |
| UL.120 | Regular Grid Sampling & Surface Tortoise Sampling | With Dummy Coordinates |
| NRW.115 | Regular Grid Sampling | Without Coordinates |
| MG.112 | Regular Grid Sampling | With Coordinates |
| SA.112 | Incomplete Regular Grid Sampling | Without Coordinates |
| G.104 | Regular Grid Sampling | With Coordinates |
| MGS.101 | Regular Grid Sampling | With Coordinates |
| CV.98 | Stratified Random Sampling | Without Coordinates |
| SC.93 | Conditioned Latin Hypercube Sampling | With Coordinates |
| BB.72 | Triangular Grid Sampling | With Coordinates |
| NRW.62 | Stratified Systematic Sampling | Without Coordinates |
| RP.62 | Regular Grid Sampling | Without Coordinates |
| SSP.58 | Stratified Random Sampling | Without Coordinates |
| NSW.52 | Random Sampling from K-Means Clustering & Stratified Random Sampling | With Coordinates |
| BB.51 | Multi Criteria Sampling | With Coordinates |
| W.50 | Conditioned Latin Hypercube Sampling | Without Coordinates |
| SC.50 | Regular Grid Sampling | With Coordinates |
| PC.45 | Stratified Systematic Sampling | Without Coordinates |
| MG.44 | Random Sampling from Regular Grid | With Coordinates |
| NRW.42 | Regular Grid Sampling | Without Coordinates |
| SM.40 | Stratified Sampling from Regular Grid | With Coordinates |
| MWP.36 | Random Sampling of Field Transects | With Coordinates |
| O.32 | Regular Grid Sampling | Without Coordinates |
| BB.30_1 | Multi Criteria Sampling | With Coordinates |
| BB.30_2 | Regular Grid Sampling | With Coordinates |

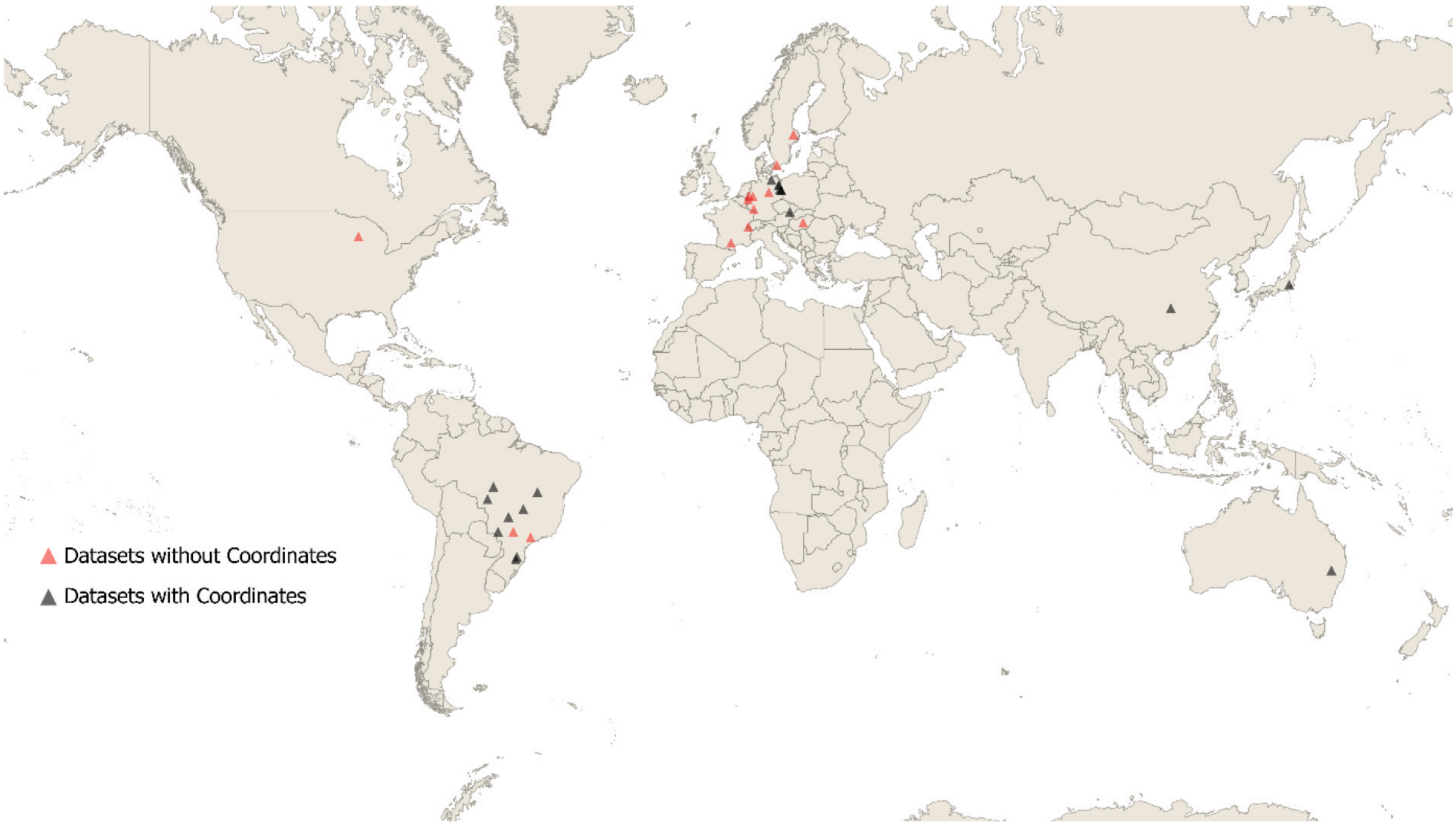

**Fig. B1.** Locations of the datasets on a global map. For datasets without coordinates, not the exact but only an approximate location is shown.

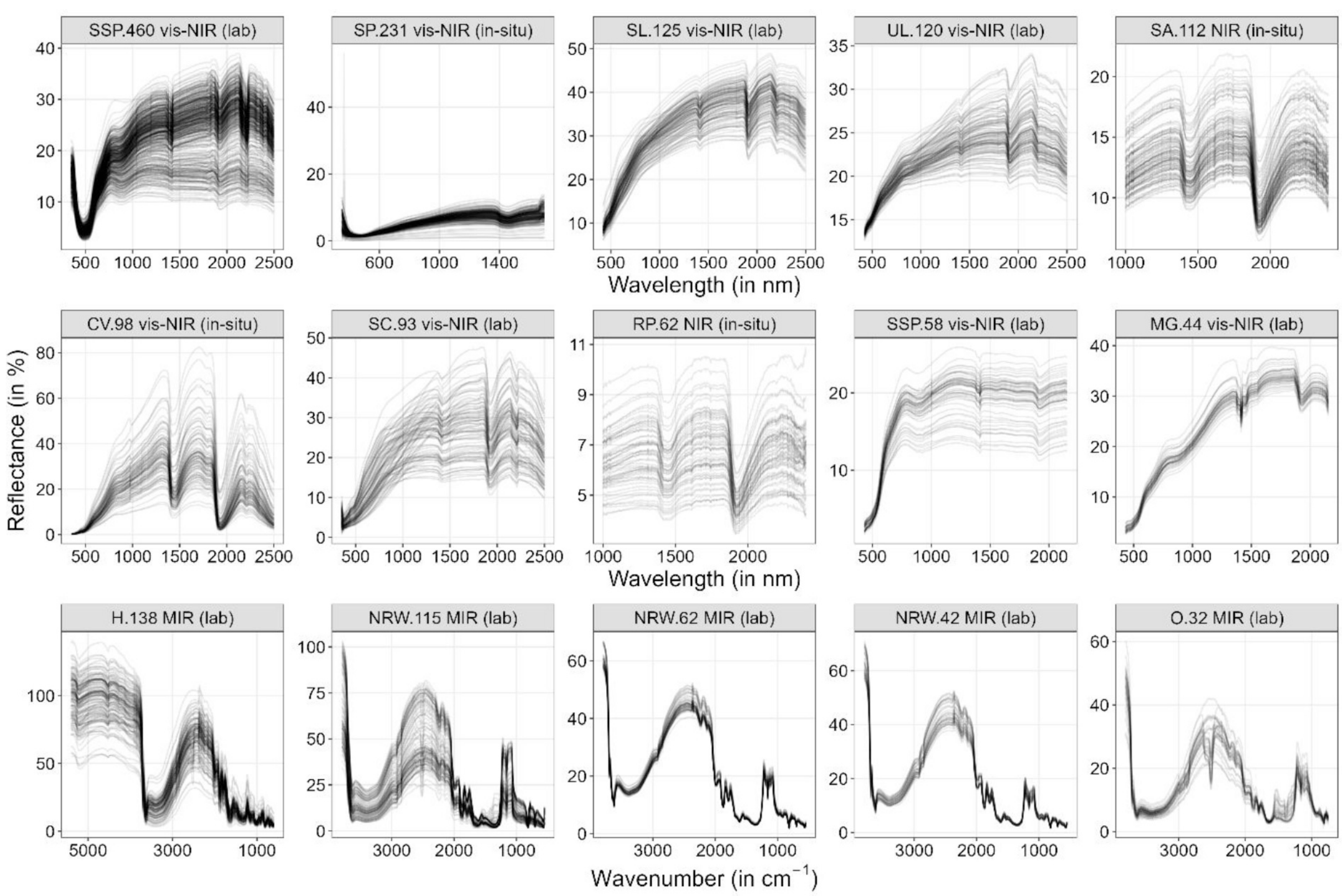

**Fig. B2.** Spectral curves from all individual samples of datasets containing vis-NIR, NIR or MIR.

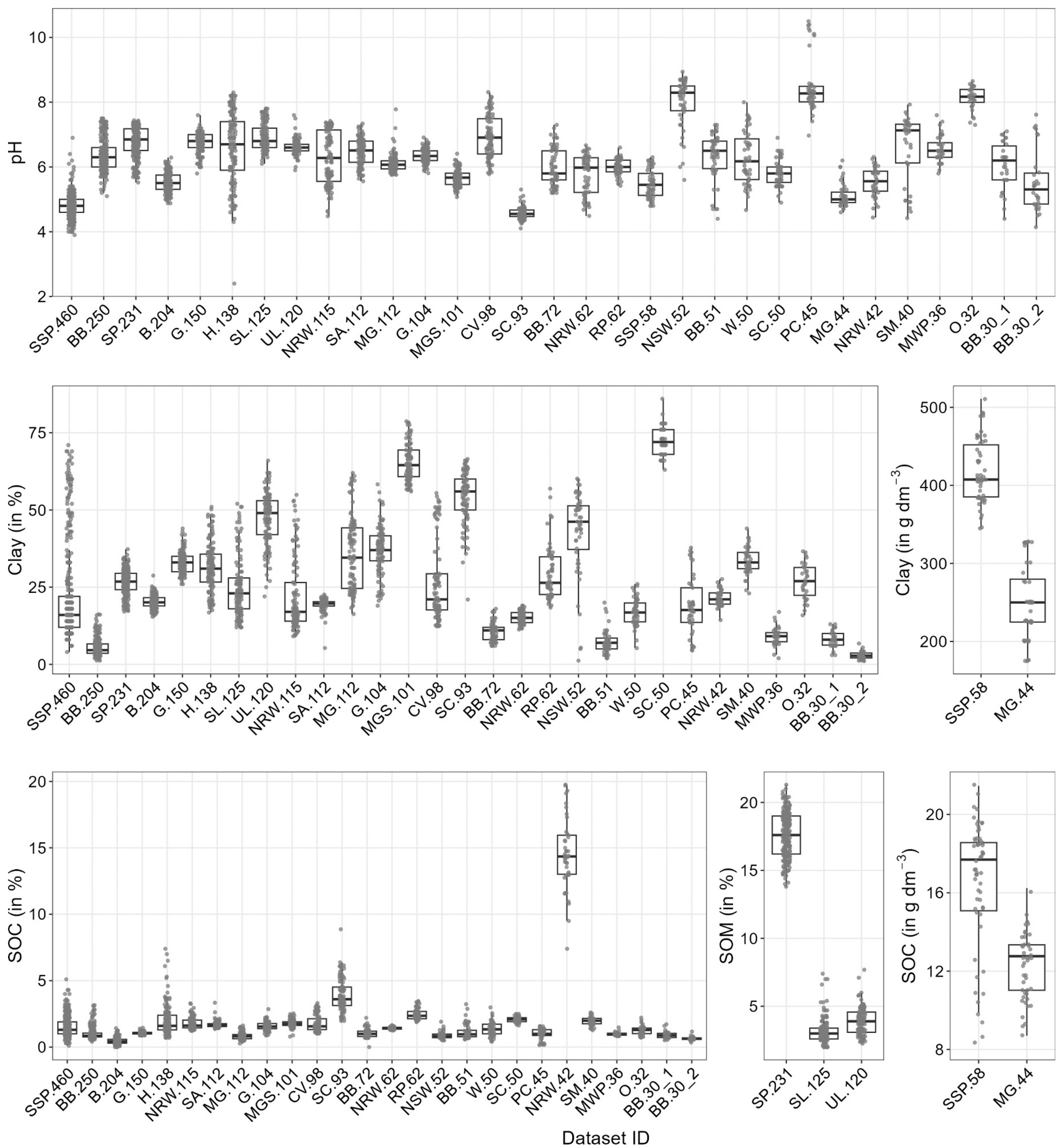

**Fig. B3.** Boxplot showing the distribution of the target soil properties. Clay and SOC are shown in different units as they were not harmonized.

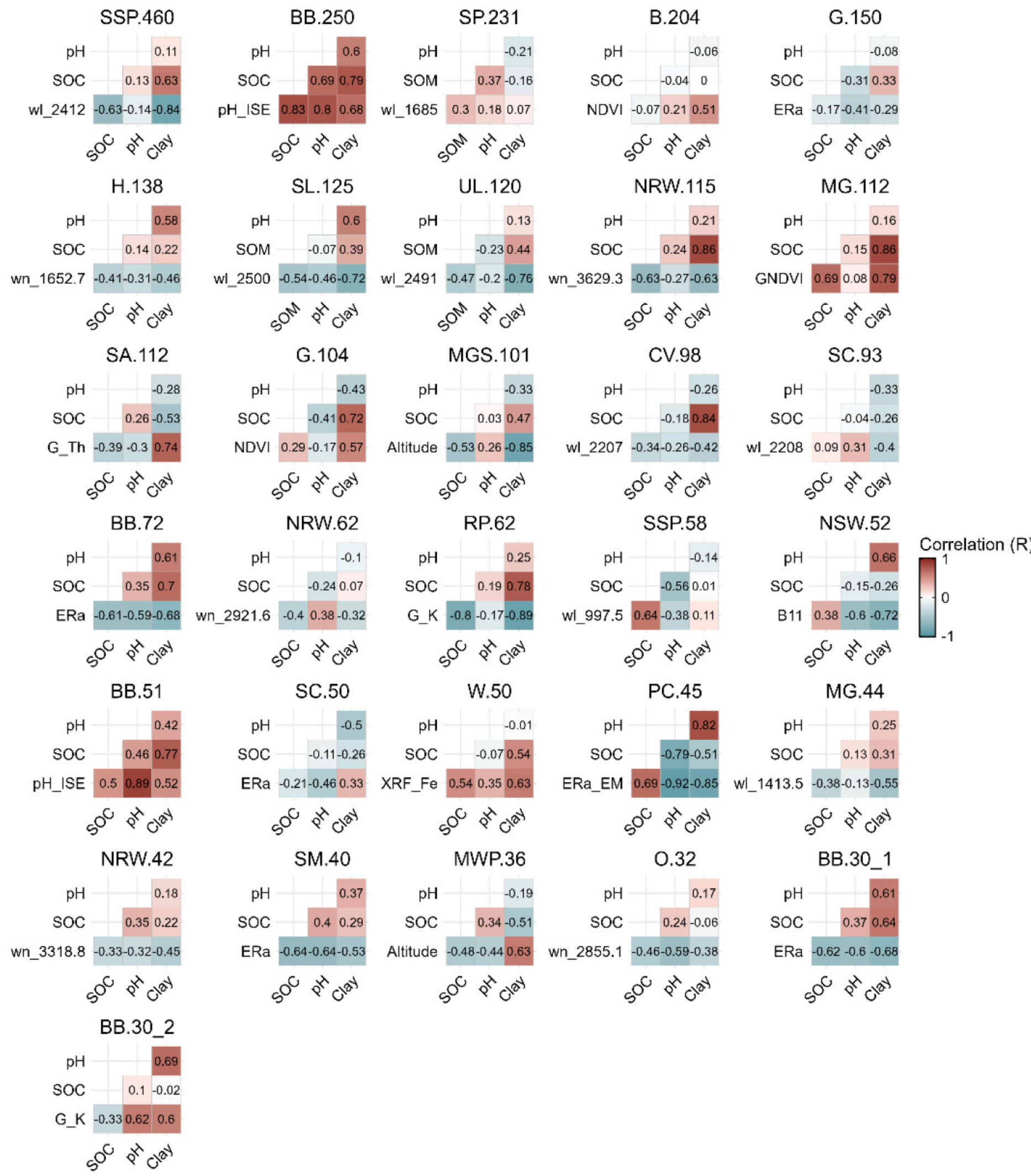

**Fig. B4.** Correlation plots for all datasets of target soil properties and a selected feature with the highest absolute correlation to the target soil properties. Correlation is given as Pearson correlation coefficient (R). Explanation of the feature codes can be looked up in the metadata of the datasets.

## Appendix C:. Further information on the benchmarking

### *Appendix C.1: Validation metrics*

RMSE and $R^2$ were determined after aggregation of fold-wise test values ($y$) and predicted values ($\hat{y}$). RMSE was calculated as:

$$RMSE = \frac{1}{n}\sum_{i=1}^{n}\sqrt{(y_i - \hat{y}_i)^2}, \tag{1}$$

where $n$ is the number of samples ($i = 1, \cdots, n$). The RMSE has an optimal value of 0 and is expressed in the unit of the target property.

For Fig. 2 and Fig. 3, we used the ordinally ranked RMSE to evaluate the performances. For every prediction task, we ranked the four learning algorithms according to the RMSE. The first rank was assigned to the learning algorithm with the lowest (i.e., best) RMSE and the fourth rank to the highest (i.e., worst) RMSE. This returns a ranking distribution of the learning algorithms, which shows how often a model performed best to worst (Fig. 2). The mean rank represents the average ordinal rank over all prediction tasks (Fig. 2 & Fig. 3). Since the magnitude of differences in RMSE between the algorithms is not measured, this approach is less prone to outliers (i.e., unusual poor or strong predictive performances).

$R^2$ was calculated as:

$$R^2 = 1 - \frac{\sum_{i=1}^{n}(y_i - \hat{y}_i)^2}{\sum_{i=1}^{n}(y_i - \bar{y})^2}, \tag{2}$$

it becomes negative if predictions are worse than the arithmetic mean of the test values $\bar{y}$ and its optimal value is 1.

### *Appendix C.2: Further results*

Fig. C1 shows which dimensionality reduction method turned out to be best within the nested CV hyperparameter selection. The best selected dimensionality reduction method was used to train the final models in the outer loop. As discussed in Section 3.2, PCA was most frequently the best method for handling the high dimensionality in vis-NIR, NIR and MIR datasets. However, with more samples, the unprocessed data (i.e., CMF = 1 in hyperparameter search) was increasingly useful for SVR. It was able to fit better models using the whole feature matrix of vis-NIR, NIR and MIR datasets with more than 100 samples (Fig. 3).

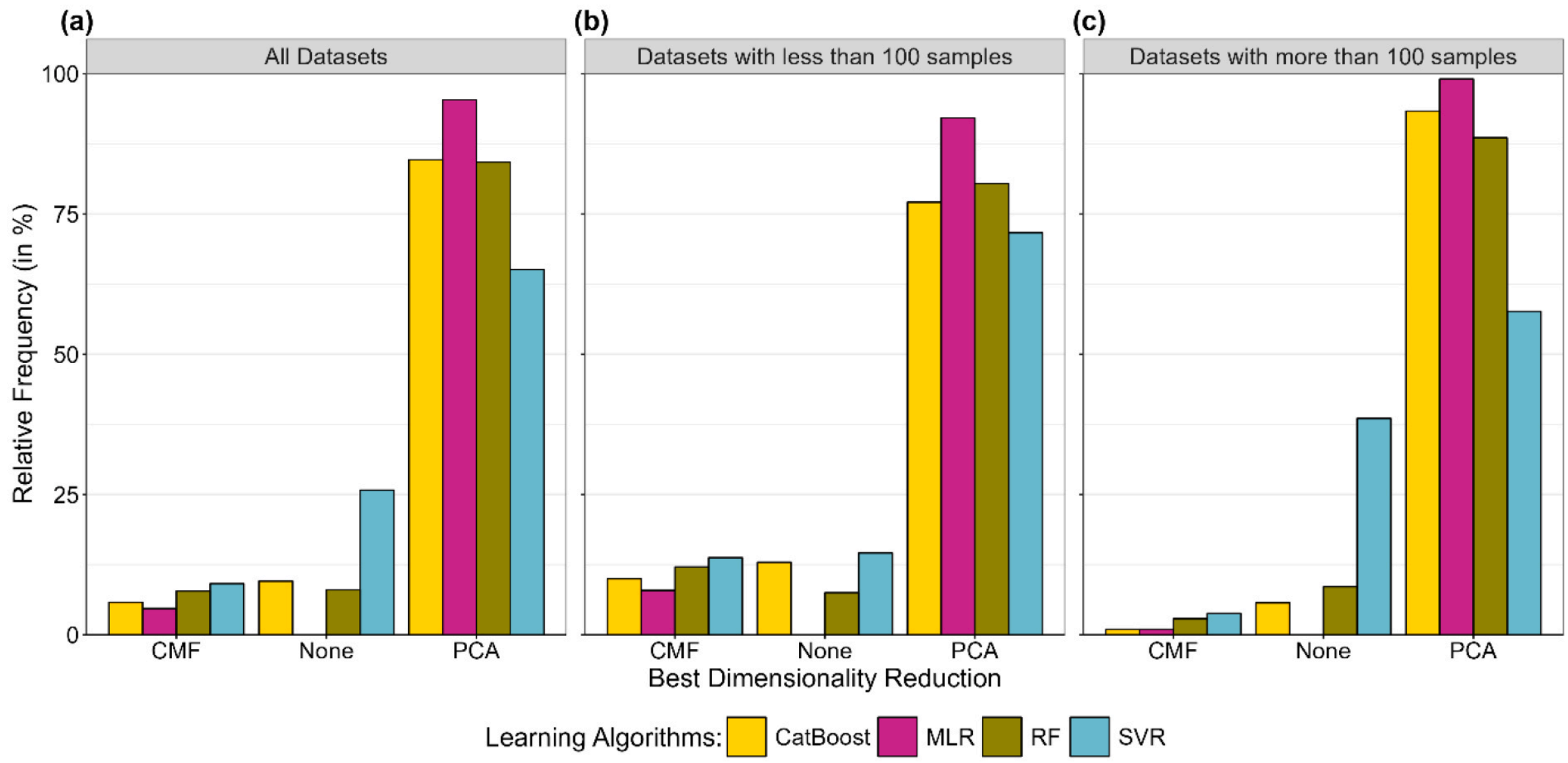

**Fig. C1.** Bar plot showing the relative frequency distribution of the best dimensionality reduction in the hyperparameter tuning for the four learning algorithms for predictions tasks from (a) all datasets, (b) datasets with more than 100 samples, and (c) datasets with less than 100 samples. "All datasets" refers here to all datasets with vis-NIR, NIR and MIR, in contrast to the previous figures, as no dimensionality reduction was used for the datasets without vis-NIR, NIR and MIR.

Fig. C2 shows the relative frequency of the best number of PCA components during the hyperparameter tuning. While there is a certain random component, it is still apparent that RF utilized less components than the other algorithms. This could explain the poor performance of RF for vis-NIR, NIR and MIR datasets. RF was not able to exploit the information encoded in the higher-order components as effectively as the other learning algorithms.

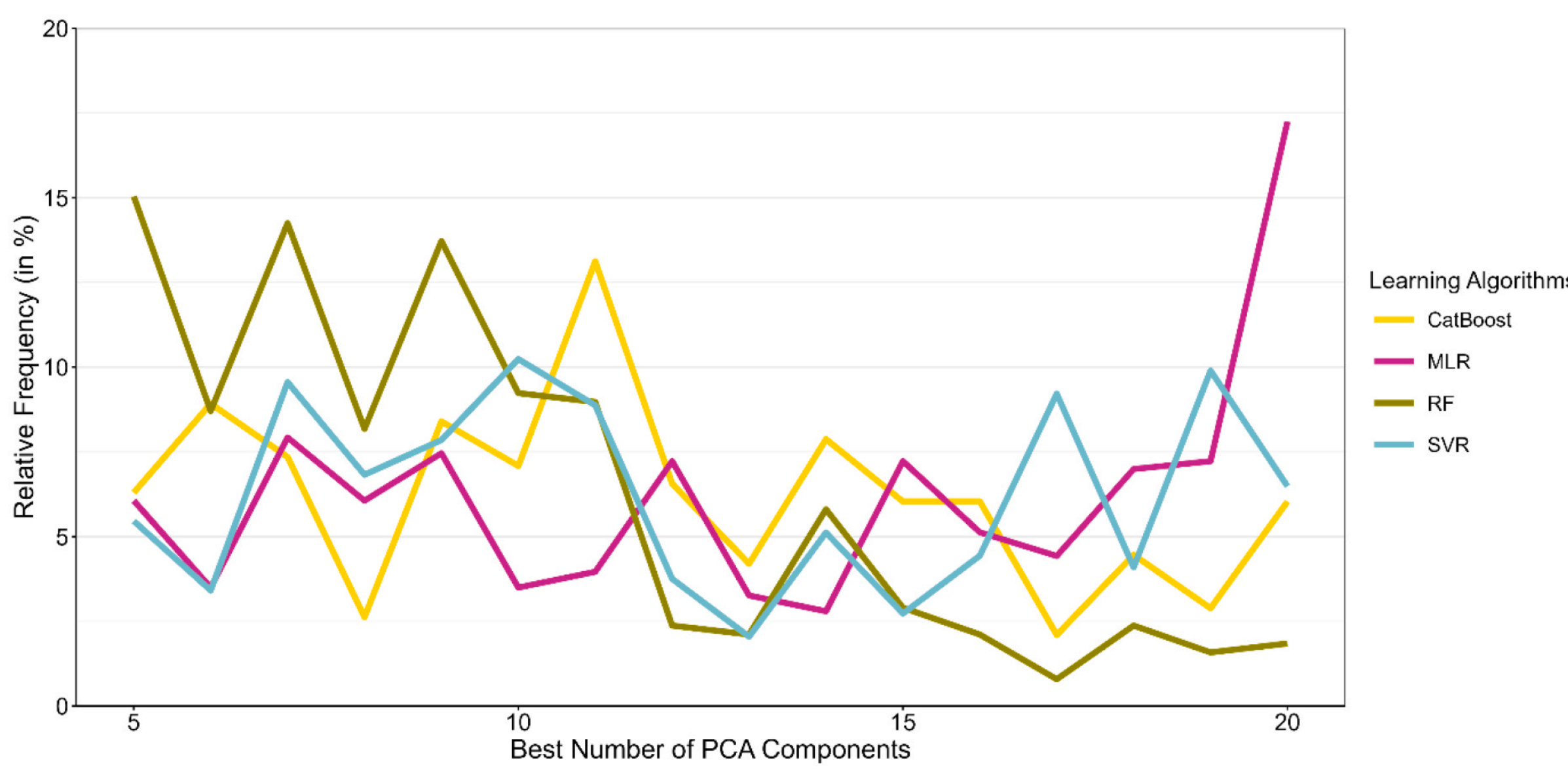

**Fig. C2.** Line plot showing the relative frequency distribution of how many PCA components were best during the hyperparameter tuning for the four learning algorithms for datasets with vis-NIR, NIR and MIR.

## Appendix D. Supplementary information

The supplementary information contains a table which shows the author contributions and the benchmarking results as $R^2$ for each prediction task, i.e., dataset and soil property. Supplementary information to this article can be found online at https://doi.org/10.1016/j.geoderma.2025.117337.

## Data availability

All datasets can be accessed through Zenodo (https://doi.org/10.5281/zenodo.14932573), an R package (github.com/JonasSchmidinger/LimeSoDa) and a Python package (https://github.com/a11to1n3/LimeSoDa). R-code and results of the benchmarking study in **Section 3** is available at https://github.com/JonasSchmidinger/LimeSoDa_benchmarking. The literature review of Appendix A is available at https://github.com/JonasSchmidinger/LimeSoDa_literature.review.